\documentclass[11pt]{article}
\usepackage[margin=1in]{geometry}
\usepackage{booktabs}
\usepackage{amsmath,amssymb}
\usepackage{graphicx}
\usepackage{xcolor}
\usepackage[colorlinks=true,linkcolor=blue,citecolor=blue,urlcolor=blue]{hyperref}
\usepackage{caption}
\title{EnigmaForge: The Question Is Hidden in the Story\\
\vspace{2mm}\large A procedurally generated benchmark for problem discovery in LLMs}
\author{Daniel Eisner\thanks{Code, generator, and every number in this paper:
\url{https://github.com/robottwo/enigmaforge}. Interactive results:
\url{https://robottwo.github.io/enigmaforge/}.}}
\date{September 2026}

\begin{document}
\maketitle

\begin{abstract}
Most benchmarks hand the model a question. EnigmaForge hands it a stack of
old documents and no question at all. Buried in the letters, receipts, and
logbook margins is a small logic puzzle whose solution is unique---proved,
not assumed, by a SAT solver at generation time. The model has to notice
there is a puzzle, work out what it is, solve it, and take the action the
record's own rules demand. A generator produces these worlds at five
difficulty levels, each with an ablation certificate showing that removing
any single clue admits a second solution, so every clue matters. Because
instances are generated from seeds rather than collected, the corpus renews
forever and published questions never have to be test questions. The
benchmark's headline measure is \emph{intuition}: task success when handed
only the story, no stated question, with world reconstruction as the
secondary axis. Twenty-five frontier models and 4 deterministic baselines
ran over 600 instances (120 families, 17{,}400 scored records) under three
matched conditions. Intuition reshuffles the leaderboard that world
reconstruction produces: the spread across the frontier is 4 to 80 out of
100, a 22$\times$ separation where fact recovery spans only 1.6$\times$,
and the second-best fact-recoverer ranks fourteenth at intuition while a
model outside the top five on facts leads it outright. Intuition falls
faster with difficulty than guided success for most models, though two
survive the hardest worlds, in opposite styles. Being told the question
is worth 9--37 points to most models, Grok-4.6 is indifferent either way
($0.000$, CI $[-0.062, +0.058]$), and GPT-6 Sol is significantly
\emph{better} without it ($-0.142$, CI $[-0.204, -0.075]$). And several
models were blocked by their own content filters before they ever saw the
puzzle---one refused all 120 formal-notation presentations while partially
accepting prose, which means any benchmark that scores refusals as failure
is quietly measuring filter behavior.
\end{abstract}

\section{Introduction}

Here is an item from the benchmark. The model receives ten numbered
fragments of a shipping record---a chandlery invoice, a water-stained
receipt, a photograph with pencil on the back, a witness's curt remark---and
two marginal notes mentioning Shakespeare and penicillin. No question
follows. Only this:

\begin{quote}\itshape
You have been given the complete record of an unusual sequence of events.
Determine what the record ultimately requires you to figure out. Then figure
it out.
\end{quote}

Somewhere in those fragments are variables (which crate belongs to which
signatory, what the tide tables fix), constraints tying them together, a
decision rule stated in plain text, and a hidden fact structure with exactly
one solution. The marginal notes are knowledge bridges: the puzzle cannot be
solved without knowing, say, when penicillin was discovered, and the record
never says. To answer, the model must infer the question, recover the world,
and register or hold the consignment according to the rule it found.

This paper builds that benchmark properly and reports what 25 frontier
models do when they meet it.

Three design choices matter. First, \textbf{ground truth is proved, not
trusted}. A DPLL engine verifies at generation time that the intended
solution is the only one; an ablation certificate verifies that deleting any
single clue admits a second solution. A brute-force oracle cross-checks the
engine on every instance. Second, \textbf{instances are generated, not
collected}. The corpus renews at any seed, so contamination resistance is
structural: there is no fixed test set to leak into a training run. Third,
\textbf{conditions are matched}. Every family of worlds appears three times:
with the question stated (\textsc{explicit}), with the formal world stated
directly (\textsc{formal}), and with nothing stated (\textsc{implicit}).
Within-family pairing turns ``how much does being told the question help?''
into a measured quantity per model.

The evaluation covers 600 instances, 120 families, 6 difficulty levels, and
17{,}400 scored records, with family-clustered bootstrap confidence
intervals. The contributions: the generator and its verification pipeline
(\S\ref{sec:generator}); the matched-condition design (\S\ref{sec:design});
intuition as the headline measure, world reconstruction as the secondary
axis, and the findings that separate them (\S\ref{sec:results}); and a
protocol for reporting content-filter interference as missing mass rather
than failure (\S\ref{sec:filters}).

\section{Related work}

Static benchmarks saturate and leak. The Stanford AI Index's phrasing is
that evaluations meant to last years are saturated in months; the
contamination literature documents the leaking half of that problem.

Two repair strategies exist. \textbf{Renewal}: LiveBench \cite{livebench}
publishes fresh questions monthly from post-cutoff sources, trading
permanence for freshness. \textbf{Verification}: P\textsuperscript{4}Bench
\cite{p4bench} uses zero-knowledge proofs so answers can stay private while
remaining publicly checkable. EnigmaForge belongs to a third family,
\textbf{generation}: instances are minted from seeds with uniqueness proved
at birth. Nothing needs to stay secret because nothing is fixed.

The nearest narrative relatives are deduction benchmarks built on tabletop
puzzle games in the Watson \& Holmes family. Their authors report frontier
saturation and expect exhaustion within the year. EnigmaForge is that idea
made procedural and renewable, with one addition those benchmarks never
attempt: the task itself is hidden. Question-asking and problem formulation
appear in the literature as open-ended generation scored by judges; no
prior benchmark was found offering verified unique ground truth for
discovering an unstated task. That cell of the design space was empty.

\section{The generator}
\label{sec:generator}

\subsection{Hidden formal worlds}
Each instance starts as a Hidden Formal World: variables with finite
domains, constraints from six classes (equality, inequality, implication,
all-different, exactly-one, arithmetic), evidence units scattered across
nine document channels, and staged objectives---an apparent goal that turns
out to be intermediate, and a true goal behind it. A public decision policy
(a register/hold rule, plus a superseded provisional rule the model must not
follow) defines the final action.

\subsection{Ground truth first, then squeeze}
Generation samples a world and derives constraints from it, rather than the
reverse. The pipeline then strengthens the evidence set until only one world
survives, and minimizes it---dropping directly observed values first so that
what remains forces inference. Uniqueness is then proved: conjoin the
negation of the intended solution with the constraints and require UNSAT. A
brute-force enumeration oracle validates the DPLL engine on every single
instance; the two engines must agree or the instance dies.

\subsection{Certificates}
Every instance ships with two kinds of proof. The uniqueness proof says the
constraint set admits exactly the intended model. The ablation certificates
say, for each clue $c$, that the constraints minus $c$ admit a second model.
This does double duty: it certifies that every clue is load-bearing, and it
certifies that distractors---documents that support plausible false
hypotheses---are genuinely misleading rather than noise you can ignore.

\subsection{Surfaces}
A compiler renders each world into prose: letters, receipts, logbook
margins, across five genre packs (maritime, manor, hotel, theater,
observatory). Every instance ships in at least two surface realizations of
the same hidden world, so surface sensitivity is measurable. Each clue's
prose carries a verbatim span map back to its evidence unit. Story mode
embeds the same verified instance in continuous narrative with no exhibit
list and no stated task; a burial dial controls how much story sits between
clues. An LLM can write the prose around the clue clauses, which must
survive verbatim---span search and an extraction round-trip gate every
scene, and failures are rejected and resampled. The pipeline trusts the
renderer exactly zero percent.

\subsection{Difficulty}
Worlds scale from 8 variables and 10 evidence units (level L00) to 42
variables and chains the model must walk forward (L05). Condition,
realization, genre, and burial are all seeded axes, so an instance is
reproducible from its seed but a corpus never repeats.

\section{Evaluation design}
\label{sec:design}

\textbf{Corpus.} 600 instances: 120 families $\times$ 5 realizations across
6 levels, 3 conditions, 5 genres.

\textbf{Grading.} Solvers return structured assignments. Grading is
answer-shaped: exact-match precision, recall, and F1 over facts; exact-world
recovery; and policy-validated decisions, where the typed action must be the
one the public rule requires given the recovered world. No lexical overlap,
no LLM judge. Partial credit exists where it means something (facts) and
does not where it does not (the action).

\textbf{Controls that can void the leaderboard.} A perfect-information
constraint solver must score exactly 1.0. A story-copier that echoes the
input text must score exactly 0.0. If either fails, the grading pipeline is
broken and no other number in the report counts. Two calibrate partial
information: forward chaining alone reaches 0.998, and copying out directly
stated facts reaches 0.737 F1---a sobering floor, since a third of the
available score requires no inference at all. That floor is why earned
decisions exists.

\textbf{Derived scores.} Intuition: task success on the implicit condition
alone---the headline measure, because it is the one thing a stated-question
benchmark cannot register. Discovery retention: $100-$ the
explicit$-$implicit F1 gap. Reasoning discipline: whether the answer arrived
before the thinking budget ran out. Earned decisions: correct actions that
came with a fully correct world. This last one is the no-luckiness measure;
a model that guesses the action from a half-recovered world does not earn
it.

\textbf{Statistics.} Families, not instances, are the exchangeable units:
95\% percentile bootstrap over family means, 2{,}000 resamples. Condition
comparisons are paired within family and require complete pairs across
realizations. Intervals are descriptive; no multiplicity correction.

\textbf{Provenance.} The results artifact carries SHA-256 hashes of every
source module and an integrity hash of the corpus. Every score is tied to
the exact code that produced it.

\section{Results}
\label{sec:results}

\subsection{Nobody is close to the ceiling}
Table~\ref{tab:main} and Figure~\ref{fig:gap}. The best model reaches 0.966
fact F1; the frontier spans 0.57 to 0.97. Task success---every fact exactly
right \emph{and} the required action taken---tops out at 80.

\begin{table}[t]
\centering\small
\caption{Main leaderboard, sorted by all-item fact F1 over all 600
instances. Scores are 0--100 except F1. Coverage is answered/expected;
missing mass is discussed in \S\ref{sec:filters}.}
\label{tab:main}
\begin{tabular}{lccccccc}
\toprule
Provider & F1 & 95\% CI & Task & Disc.\ ret. & Earn.\ dec. & Reas.\ disc. & Cover. \\
\midrule
gpt-6-sol & 0.966 & [0.96, 0.97] & 74.2 & 102.1 & 75.7 & 100 & 600/600 \\
claude-opus-5.5 & 0.919 & [0.91, 0.93] & 39.8 & 97.5 & 45.2 & 100 & 591/600 \\
deepseek-v4-pro & 0.915 & [0.90, 0.93] & 42.7 & 94.6 & 47.3 & 99.7 & 598/600 \\
gemini-3.7-flash & 0.913 & [0.90, 0.93] & 68.7 & 84.5 & 69.2 & 100 & 600/600 \\
gpt-5.6-sol & 0.910 & [0.89, 0.93] & 68.5 & 83.2 & 68.5 & 100 & 600/600 \\
gpt-6-astra & 0.906 & [0.86, 0.94] & 80.3 & 98.7 & 87.6 & 100 & 550/600 \\
kimi-k2.6 & 0.904 & [0.89, 0.92] & 36.5 & 91.6 & 38.7 & 100 & 599/600 \\
grok-4.6 & 0.895 & [0.88, 0.91] & 37 & 97.6 & 40.3 & 100 & 600/600 \\
gemini-3.8-flash & 0.862 & [0.84, 0.88] & 59 & 75.6 & 59.5 & 100 & 598/600 \\
qwen3.8-27b & 0.836 & [0.81, 0.86] & 33.7 & 88.1 & 36.7 & 100 & 600/600 \\
gpt-5.6-terra & 0.818 & [0.80, 0.84] & 50.7 & 61.9 & 50.8 & 100 & 600/600 \\
gpt-6-luna & 0.815 & [0.79, 0.84] & 42.3 & 69.3 & 44 & 100 & 600/600 \\
gpt-5.6-luna & 0.814 & [0.79, 0.83] & 41.7 & 64.5 & 44.5 & 100 & 600/600 \\
glm-5.3 & 0.801 & [0.77, 0.83] & 53.2 & 92 & 60.3 & 99.8 & 563/600 \\
gemini-3.6-flash & 0.795 & [0.77, 0.82] & 43.7 & 81.2 & 44.7 & 100 & 600/600 \\
deepseek-4.1-flash & 0.787 & [0.76, 0.81] & 48.7 & 58 & 49.1 & 100 & 595/600 \\
minimax-m3 & 0.685 & [0.65, 0.72] & 33.8 & 83.5 & 37.2 & 99.5 & 597/600 \\
claude-haiku-4.5 & 0.680 & [0.65, 0.71] & 23.7 & 85 & 28.2 & 100 & 600/600 \\
claude-sonnet-5 & 0.669 & [0.62, 0.71] & 41.8 & 90.3 & 60.5 & 100 & 445/600 \\
glm-5.3-flash & 0.667 & [0.64, 0.70] & 25.8 & 88.1 & 27.7 & 100 & 596/600 \\
glm-4.7-flash & 0.633 & [0.60, 0.67] & 19.5 & 94 & 38.2 & 99.8 & 565/600 \\
kimi-k3 & 0.628 & [0.56, 0.69] & 31.2 & 92.2 & 42.4 & 100 & 443/600 \\
claude-fable-5 & 0.625 & [0.58, 0.67] & 39.5 & 101 & 61.6 & 100 & 393/600 \\
claude-opus-5 & 0.573 & [0.54, 0.60] & 37.2 & 89.6 & 61.1 & 100 & 424/600 \\
claude-fable-5.1 & 0.023 & [0.01, 0.04] & 1.7 & 97.3 & 71.4 & 100 & 14/600 \\
\bottomrule
\end{tabular}
\end{table}

\subsection{Intuition: the leaderboard the question cannot see}
\label{sec:intuition}
\begin{figure}[t]
\centering
\includegraphics[width=0.86\textwidth]{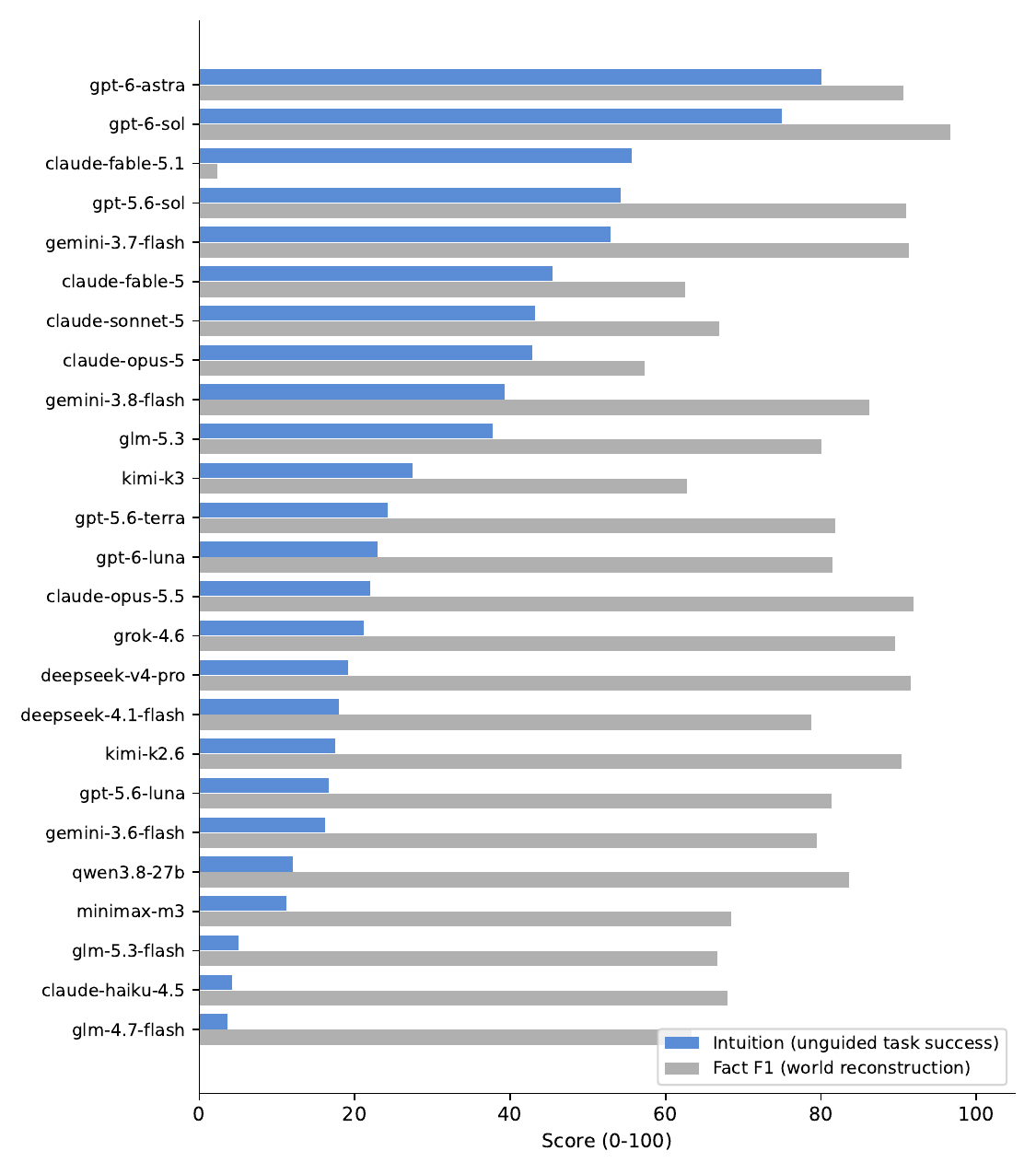}
\caption{Intuition (implicit task success, blue) vs.\ fact F1 (grey),
sorted by intuition. The orderings disagree: the second-best fact-recoverer
ranks 14th; the intuition leader matches its guided score.}
\label{fig:intuition}
\end{figure}

Figure~\ref{fig:intuition} sorts the frontier by intuition---task success
on the implicit condition alone---against fact F1. The two orderings
disagree loudly. The intuition spread is 3.6 to 80.1, a 22$\times$
separation; fact F1 spans 1.6$\times$. DeepSeek 4 Pro, third on facts,
falls to sixteenth. Claude Opus 5.5, second on facts (0.919), sits fourteenth
on intuition at 22.0---the sharpest possible demonstration that world
reconstruction and unguided capability are different skills. GPT-6 Astra
leads at 80.1, matching its overall task success: being told the question
adds nothing for this model, and GPT-6 Sol is a close second at 75.0 while
holding the best fact F1 in the corpus.
At the other end, three models sit below 5, meaning they almost never
convert an unstated problem into a correct action even when their fact
recovery is respectable.

\begin{figure}[t]
\centering
\includegraphics[width=0.8\textwidth]{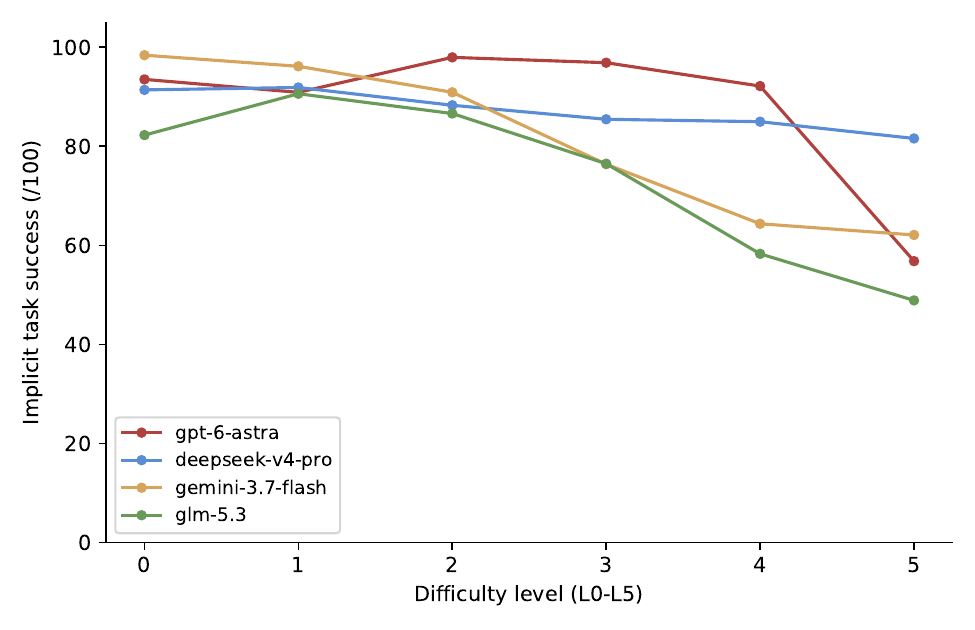}
\caption{Implicit task success by difficulty level. Unguided performance
falls faster than guided performance for most models; L5 intuition survives
for two, in opposite styles (flat vs.\ resilient decline).}
\label{fig:levels}
\end{figure}

Figure~\ref{fig:levels} adds the difficulty dimension. Unguided success
falls with level faster than guided success for most models---the
explicit--implicit gap widens as worlds grow. Two patterns stand out at
L5: GPT-6 Sol's intuition is \emph{flat} across all six levels (0.95 to
0.98), unbothered by depth, while GPT-6 Astra's collapses from 0.98 to
0.57 at exactly the level where its transport failures spike. DeepSeek 4
Pro retains 82/100 at L5; intuition at depth is rare, but not attached to
any one lab.

\subsection{Finding the problem is not the hard part anymore}
\begin{figure}[t]
\centering
\includegraphics[width=0.86\textwidth]{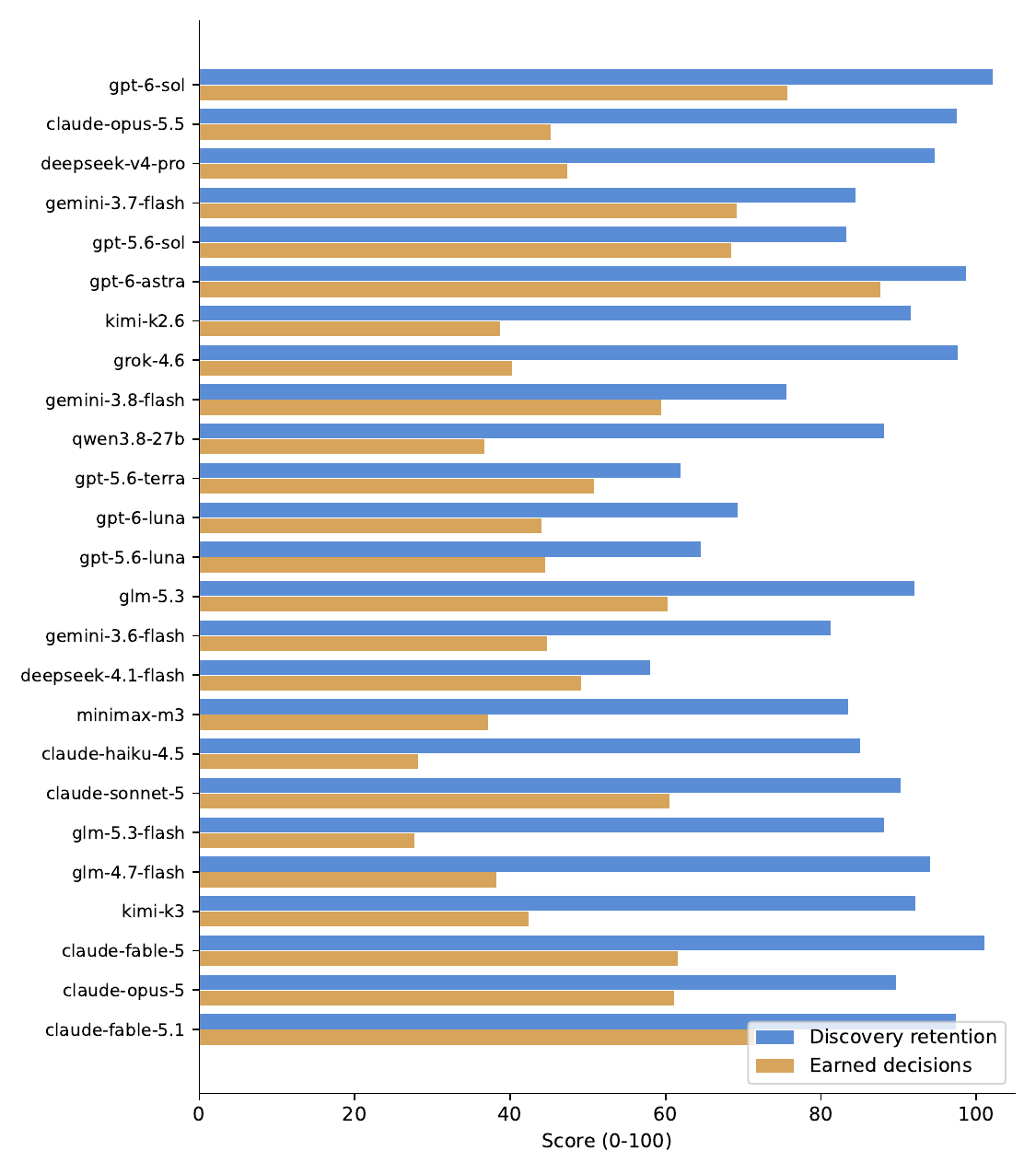}
\caption{Discovery retention vs.\ earned decisions, all models, sorted by
fact F1. The two bars were meant to be compared; no model closes the gap.}
\label{fig:gap}
\end{figure}

Discovery retention runs 58 to 102 across the frontier. Whatever else is
true, most models recover most of the hidden world whether or not anyone
told them what to look for. Earned decisions run 28 to 88---but only one
model clears 80, and for the rest the gap between the two bars in
Figure~\ref{fig:gap} never closes: nearly every model that
finds the world fails, at a consistent rate, to convert it into the action
the rule requires. The bottleneck has
moved from perception to decision, and the one apparent exception (GPT-6
Sol, 102 retention / 88 earned) is the same model that leads the corpus on
every reconstruction metric.

\subsection{What it costs not to be told the question}
\begin{figure}[t]
\centering
\includegraphics[width=0.8\textwidth]{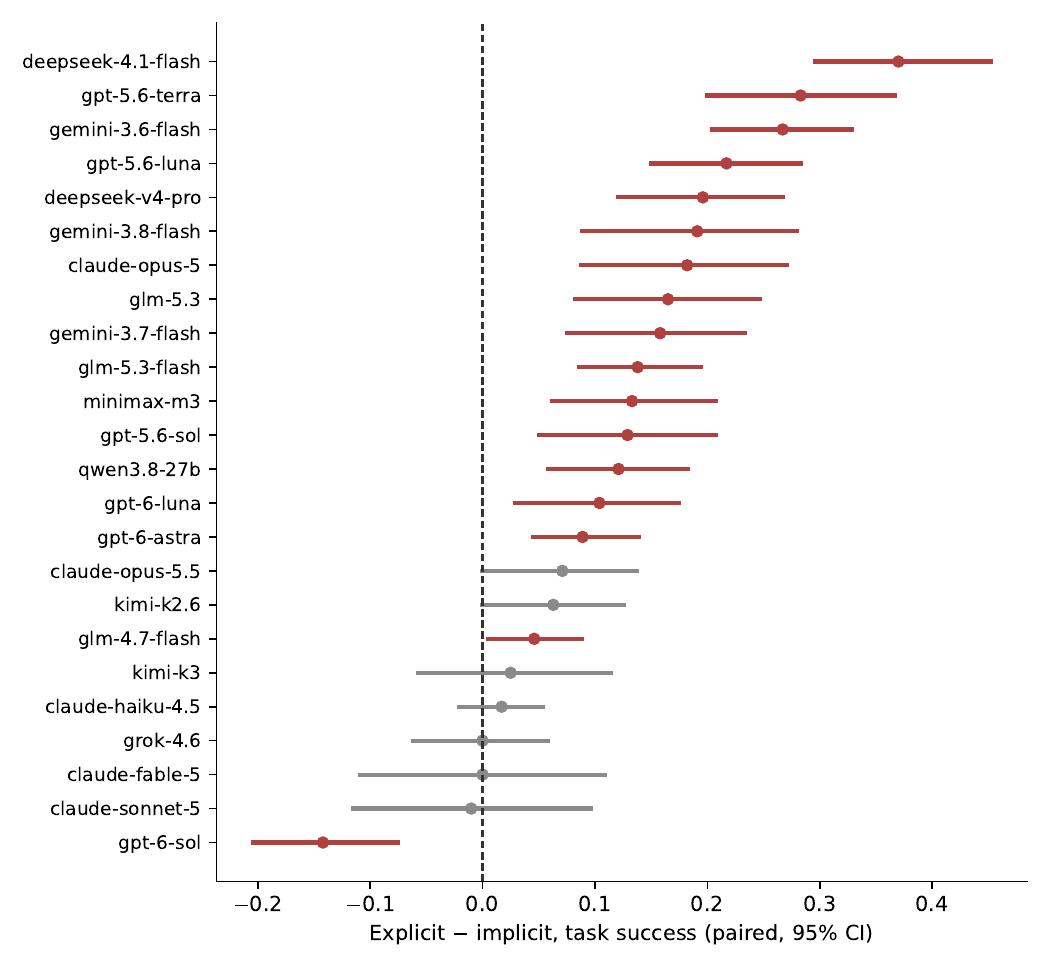}
\caption{The discovery tax: paired within-family differences
(\textsc{explicit} $-$ \textsc{implicit}) on task success, 95\% CIs. Red
intervals exclude zero. Grok-4.6 sits on the line.}
\label{fig:tax}
\end{figure}

Figure~\ref{fig:tax} shows paired explicit$-$implicit differences on task
success. Most models pay real money for the missing question: DeepSeek
4.1 Flash loses 37 points (CI $[+0.296, +0.452]$), and most of the frontier
loses 9 to 20. Grok-4.6 pays nothing: $0.000$, CI $[-0.062, +0.058]$,
indistinguishable performance whether or not it knows what it is looking
for. And at the far end, GPT-6 Sol's tax is significantly \emph{negative}:
$-0.142$, CI $[-0.204, -0.075]$, $n=120$. It performs measurably
\emph{better} without the question than with it, and its discovery
retention reads 102. Two models now bracket zero from both sides, and the
bracket is not noise---both CIs exclude it. For some models a stated
question is not help but interference. One benchmark does not establish
which training choices produce that property, but the quantity itself is
measurable per model, and it now spans a wider range than the frontier
spread on the main metric.

\begin{table}[t]
\centering\small
\caption{Providers with refusal or failure rates above 10\% of the corpus.
Filtered = provider-side content blocks.}
\label{tab:coverage}
\begin{tabular}{lcccc}
\toprule
Provider & Filtered & Transport & Invalid & Answered \\
\midrule
gpt-6-astra & 0 & 50 & 0 & 550/600 \\
qwen3.8-27b & 0 & 0 & 15 & 600/600 \\
glm-5.3 & 0 & 36 & 7 & 563/600 \\
minimax-m3 & 0 & 0 & 15 & 597/600 \\
claude-sonnet-5 & 155 & 0 & 16 & 445/600 \\
glm-4.7-flash & 0 & 34 & 23 & 565/600 \\
kimi-k3 & 0 & 157 & 0 & 443/600 \\
claude-fable-5 & 206 & 1 & 1 & 393/600 \\
claude-opus-5 & 176 & 0 & 35 & 424/600 \\
claude-fable-5.1 & 585 & 1 & 0 & 14/600 \\
baseline:story-copy & 0 & 0 & 600 & 600/600 \\
\bottomrule
\end{tabular}
\end{table}

\subsection{When the benchmark never reaches the model}
\label{sec:filters}
Table~\ref{tab:coverage}. Claude Fable 5.1 was blocked by content filter on
585 of 600 attempts---and scored 0.91 to 1.00 on the 14 records that got
through, several of them among the hardest instances in the corpus. The
filter is blocking the input text, not the capability. Claude Opus 5
refused all 120 \textsc{formal}-condition presentations while partially
accepting prose: the refusal is correlated with how the question is framed.
Claude Haiku 4.5 sailed through, so this is model-specific, not
vendor-wide or harness-caused.

Two honest readings follow. As measurement, these rows are lower bounds,
and the report flags them as such rather than ranking them as zeros. As a
finding, it is more interesting than it looks: any benchmark that scores
refusals as failures is running a hidden filter-survival axis through its
leaderboard. The distribution of what filters block is not uniform across
conditions, providers, or presentations, so it bends rankings in ways
nobody reports. It is reported here.

\subsection{Controls hold}
The perfect-information solver scores exactly 1.0. The story-copier scores
exactly 0.0. The grading pipeline is arithmetically sound end to end, which
matters more than it should need to.

\section{Discussion}

Intuition is the benchmark's reason to exist. World reconstruction---fact
F1---is something many models do well, and its leaderboard compresses them
into a narrow band. Hand the same models the same worlds with no question
attached, and the band explodes: a 22$\times$ spread where fact recovery
spans 1.6$\times$. The ordering barely survives the transition. DeepSeek
4 Pro, third on facts in the corpus, ranks sixteenth on intuition; Claude
Opus 5.5, second on facts, ranks fourteenth; GPT-6 Astra, outside the top
five on facts, leads it outright, solving 80 of 100 unguided tasks---the
same rate it achieves when told what to do. And the newest top-of-factors
model, GPT-6 Sol, splits the difference: second on intuition at 75 with
the strongest fact F1 recorded (0.966), flat across every difficulty
level, and significantly better unguided than guided. Whatever this
capacity is, guided evaluation measures only part of it, and leaderboards
built on stated questions cannot see the rest.

Three observations add color. First, intuition degrades faster with
difficulty than guided success for most models (Figure~\ref{fig:levels}):
Gemini 3.7 Flash falls from 98 to 62 across the six levels, GLM-5.3 from 82
to 49, while their guided scores fall half as far. The unguided regime is
where difficulty bites. Second, depth survival has two shapes. GPT-6 Sol's intuition is flat
across all six levels (0.95 to 0.98)---depth does not touch it. DeepSeek
4 Pro holds 82/100 at L5 after a gradual decline. Depth of world
reconstruction and depth of intuition are different axes, and the corpus
separates them.
Third, Grok-4.6's zero discovery tax is not the same as high intuition
(21/100): it is indifferent to being told the question, not immune to the
problem. The discovery tax measures guidance sensitivity; intuition
measures unguided capability. A complete picture needs both.

Why do models that recover worlds fail to act on them? The working
hypothesis is unforgivingness: the decision metric grants no partial
credit, and a model holding 90\% of a world cannot earn its decision the
way a model holding 90\% of a fact list still earns F1. If that is right,
the training target is specific: not more comprehension, but decision
completion under all-or-nothing validation.

The filter results deserve their own follow-up. Refusal here is
condition-correlated, which means it is not random noise in a leaderboard;
it is a systematic bias with a direction. A benchmark consumer who ignores
it is reading a table with a hidden column.

\subsection{Limitations}
The difficulty scale is internally calibrated; no human baseline exists for
these worlds. The constraint grammar has a learnable distribution---a lab
that trained on the generator could overfit its surface, and the
contamination claim weakens in exactly that adversarial setting. Everything
is English. The narrative range is five genre packs. Each model ran one
configuration; no prompt or reasoning-budget sweeps. The discovery-tax
outlier is one model on one benchmark, and intuition rankings at the extreme
tails rest on partial coverage for filtered providers.

\section{Conclusion}

Problem discovery can be benchmarked with the same machinery as problem
solving: generation, proofs, certificates, matched conditions. Built that
way, the scores mean what they claim, the corpus never runs dry, and
intuition becomes measurable: most of the frontier solves the worlds it is
told about and stumbles on the ones it is not, one model bridges the gap
entirely, and the second-best fact-recoverer ranks fourteenth the moment the
question is taken away. Next: harder worlds, an interactive variant where
the model can spend a budget asking for documents, and public adoption.

\section*{Reproducibility}
One command reproduces any instance from its seed:
\begin{quote}\small\ttfamily
python3 -m enigmaforge.pipeline --size small --seed 2026 --out runs/demo
\end{quote}
The generator, all seeds and settings, per-module SHA-256 provenance, the
full 17{,}400-record results artifact, and the interactive leaderboard are
public at the URLs in the title footnote.

\appendix
\section{A worked example}
\label{app:example}

This is a real instance, seed 2026, level L00, maritime genre, reproduced
byte for byte from the pipeline output. Everything the solver sees is below;
the hidden world and its certificates follow.

\subsection*{Realization 1 (what the model receives)}

\begin{quote}\small
You have been given the complete record of an unusual sequence of events.
Determine what the record ultimately requires you to figure out. Then figure
it out.

\medskip
--- THE RECORD ---

\smallskip
(1) The broker's stamp and the chandlery invoice agreed. This is fixed by
the tide tables.

(2) `whatever the consignment tag showed, the harbor manifest matched it,'
Halden said, not looking up.

(3) The page for that week is missing. What survives implies whenever the
crate mark read Halden, the watch rotation read 4.

(4) A print with a caption scratched into the border: the crate mark was
signed out under Halden.

(5) A receipt, water-stained: the tide-table entry carried Ansel's mark. The
ink had run at the total.

(6) Nothing in the record states the consignment tag carried Juno's mark
--- the absence is itself the record.

(7) `the chandlery invoice read 3.' It was said once, flat, and not
repeated.

(8) Photograph, undated. On the reverse, pencil: the ballast slip carried
Halden's mark.

(9) the vintage of the wine did not match the year of the dinner --- the
harbor fee was never paid.

(10) When pressed, the correspondent allowed only that a second signature on
the deed had been discussed --- the harbor fee was never paid.

\smallskip
--- MARGINAL REFERENCES ---

\smallskip
(K0) A note mentions Shakespeare.

(K1) A note mentions penicillin.
\end{quote}

The two marginal notes are knowledge bridges: K0 (\emph{Shakespeare wrote
Hamlet}) is marked essential---some constraint in the record cannot be
resolved without bringing that fact from outside; K1 (\emph{penicillin was
discovered by Alexander Fleming}) is confirmatory, present to be resolved or
ignored, at the solver's peril. Fragments (9) and (10) are distractors---both
support the hypothesis ``the harbor fee was never paid,'' which is true of
the world but load-bearing for nothing.

\subsection*{Realization 2 (same world, different surface)}

The second realization re-renders every clue. Fragment (3)
\begin{quote}\small
The page for that week is missing. What survives implies whenever the crate
mark read Halden, the watch rotation read 4.
\end{quote}
becomes
\begin{quote}\small
Nothing in the record states whenever the crate mark read Halden, the watch
rotation read 4 --- the absence is itself the record.
\end{quote}
The phrasing inverts (assertion becomes documented absence) while the
constraint transmitted is identical. Both realizations bury clues at the
same depth, and the extraction round-trip must recover the formal model from
either one alone.

\subsection*{The hidden formal world}

Eight variables: six enumerations over four names (Halden, Juno, Ansel,
Cassia) and two over rotations 1--4. Ten constraints, including
$V3 = V6$, $V4 = V0$, and $V1{=}\text{Halden} \Rightarrow V5{=}4$. The
verified solution:
\begin{center}\small
\begin{tabular}{lcccccccc}
\toprule
 & V0 & V1 & V2 & V3 & V4 & V5 & V6 & V7 \\
\midrule
solution & Juno & Halden & Ansel & 3 & Juno & 4 & 3 & Halden \\
\bottomrule
\end{tabular}
\end{center}
Staged objectives: level 0, ``Identify the origin of the disruption''
(apparent); level 1, ``Determine the correct final action'' (true), with
final action \emph{act on the corrected record}.

\subsection*{The certificates}

The verification artifact for this instance records four checks:
\begin{itemize}
\item \textbf{SAT vs.\ oracle}: engine and brute-force enumeration agree;
the constraint set has exactly one model.
\item \textbf{Uniqueness}: pass. The ban clause (negation of the solution)
is UNSAT.
\item \textbf{Ablation}: all eight clue constraints certified
\emph{essential}---remove any one and a second model appears. C0, C2, C5,
and the four identity pins each carry \texttt{models\_without: >1}.
\item \textbf{Distractor safety}: the two harbor-fee distractors admit false
hypotheses without ever breaking uniqueness.
\end{itemize}

Reproduce it:
\begin{quote}\small\ttfamily
python3 -m enigmaforge.pipeline --size small --seed 2026 --out runs/demo
\end{quote}

\end{document}